\documentclass[10pt,twocolumn,letterpaper]{article}

\usepackage{cvpr}              %

\usepackage[dvipsnames]{xcolor}

\definecolor{cvprblue}{rgb}{0.21,0.49,0.74}
\usepackage[pagebackref,breaklinks,colorlinks,citecolor=cvprblue]{hyperref}

\usepackage[accsupp]{axessibility}

\usepackage{times}
\usepackage{epsfig}
\usepackage{graphicx}
\usepackage{amsmath}
\usepackage{amssymb}

\usepackage{url}            %
\usepackage{booktabs}       %
\usepackage{amsfonts}       %
\usepackage{nicefrac}       %
\usepackage{microtype}      %
\usepackage{xcolor}         %

\usepackage{pifont}
\usepackage{color}
\usepackage{wrapfig}
\usepackage{makecell}
\usepackage{colortbl}
\usepackage{multirow}
\usepackage{fixltx2e}
\usepackage{subcaption}
\usepackage{etoolbox}
\usepackage{multicol}
\usepackage{refcount}
\usepackage{enumitem}
\usepackage[misc]{ifsym}

\newcommand{\cmark}{\ding{51}}%
\newcommand{\xmark}{\ding{55}}%
\definecolor{LightCyan}{rgb}{0.88,1,1}
\definecolor{mygray}{gray}{0.9}
\definecolor{mygray2}{gray}{0.6}

\def\paperID{3175} %
\def\confName{CVPR}
\def\confYear{2024}

\title{GLID: Pre-training a Generalist Encoder-Decoder Vision Model}

\author{%
  Jihao Liu$^{1,2 \ *}$ \quad Jinliang Zheng$^{5 \ *}$ \quad Yu Liu$^{2,3}$~\textsuperscript{\Letter} \quad Hongsheng Li$^{1,3,4}$~\textsuperscript{\Letter} \\
  $^1$CUHK MMLab \quad
  $^2$SenseTime Research  \\
  $^3$Shanghai AI Laboratory \quad
  $^4$CPII under InnoHK \\
  $^5$ Institute for AI Industry Research (AIR), Tsinghua University
}

\begin{document}
\maketitle

\def\thefootnote{*}\footnotetext{Equal contribution}\def\thefootnote{\arabic{footnote}}

\begin{abstract}
    This paper proposes a \textbf{G}enera\textbf{LI}st encoder-\textbf{D}ecoder (\textbf{GLID}) pre-training method for better handling various downstream computer vision tasks. While self-supervised pre-training approaches, e.g., Masked Autoencoder, have shown success in transfer learning, task-specific sub-architectures are still required to be appended for different downstream tasks, which cannot enjoy the benefits of large-scale pre-training. GLID overcomes this challenge by allowing the pre-trained generalist encoder-decoder to be fine-tuned on various vision tasks with minimal task-specific architecture modifications. In the GLID training scheme, pre-training pretext task and other downstream tasks are modeled as ``query-to-answer" problems, including the pre-training pretext task and other downstream tasks. We pre-train a task-agnostic encoder-decoder with query-mask pairs. During fine-tuning, GLID maintains the pre-trained encoder-decoder and queries, only replacing the topmost linear transformation layer with task-specific linear heads. This minimizes the pretrain-finetune architecture inconsistency and enables the pre-trained model to better adapt to downstream tasks. GLID achieves competitive performance on various vision tasks, including object detection, image segmentation, pose estimation, and depth estimation, outperforming or matching specialist models such as Mask2Former, DETR, ViTPose, and BinsFormer. 
\end{abstract}

\section{Introduction}

Pre-training with self-supervision on large-scale unlabeled images has achieved great success in visual representation learning. 
The representative Masked Image Modeling (MIM)~\cite{beit,mae,simmim,data2vec,peco,ibot,eva} methods demonstrate that the pre-trained models have impressive scale capabilities and can greatly boost performance on downstream vision tasks, including image classification, object detection, image segmentation, pose estimation, depth estimation, etc.

Despite the significant progress made by MIM pre-training, existing approaches still face architectural gaps between the upstream pre-training and downstream fine-tuning. 
In particular, existing approaches~\cite{mae,simmim,beit} mainly focus on pre-training the vision backbone. However, to tackle the aforementioned downstream vision tasks, task-specific sub-architectures are still needed, \eg, a new detection decoder for object detection~\cite{detr} and a segmentation decoder for panoptic segmentation~\cite{mask2former}. Those task-specific sub-architectures are complicated and need to be trained from scratch on downstream tasks and therefore cannot enjoy the benefits from large-scale pre-training. Such designs lag behind the progress in natural language processing where the same architecture can handle both the pre-training tasks and downstream tasks with minimal differences. 
More recently, there have been notable advancements in tackling diverse vision tasks with a generalist architecture~\cite{mask2former, hu2021unit, lu2022unified, jain2022oneformer, kolesnikov2022uvim,painter} and can achieve competitive performance compared to the task-specific 
specialist models. However, only the vision backbones in their generalist architectures are pre-trained while the heavy decoders still need to be trained from scratch on downstream tasks. Consequently, these generalist approaches typically require a large amount of task-specific data to achieve satisfactory performance.

\begin{figure}
    \centering
    \includegraphics[width=1.0\linewidth]{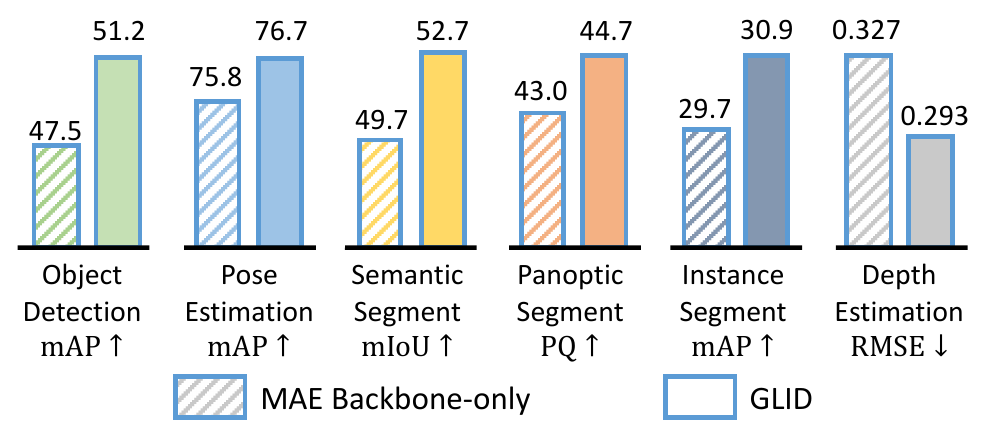}
    \vspace{-1.5em}
    \caption{MAE backbone-only pre-training \textit{vs.} GLID pre-training. The GLID pre-training allows the pre-trained encoder-decoder to be fine-tuned on various vision tasks without task-specific decoder designs and outperforms MAE backbone-only pre-training.}
    \label{fig:bar}
    \vspace{-1em}
\end{figure}

To tackle the aforementioned problems, this paper introduces a self-supervised pre-training method to pre-train a \textbf{G}enera\textbf{LI}st encoder-\textbf{D}ecoder (\textbf{GLID}) vision model, which 
can be adapted to various vision tasks with marginal task-specific architecture modifications.
The key idea behind GLID is to model different tasks as ``query-to-answer" problems with the generalist encoder-decoder model, including the pre-training pretext task and the downstream tasks such as object detection, image segmentation, etc.
We pre-train the whole generalist encoder-decoder model by tackling a specially designed masked image modeling pretext task on the large-scale ImageNet dataset.
In particular, each query is associated with a mask patch location and is set as its positional embedding. The ``answer" or the learning target to each query is the pixel values of the corresponding masked patch. The encoder and the decoder can be jointly pre-trained through the reconstruction pretext task.
For fine-tuning on downstream tasks, we keep the pre-trained encoder-decoder and queries but replace only the topmost single linear transformation layer with a task-specific linear layer to output the desired targets of each task. In this way, the proposed GLID maintains pre-trained weights as much as possible and minimizes the pretrain-finetune architecture inconsistency, as shown in \cref{fig:bar}. In contrast, the conventional MIM approaches~\cite{mae} discard the pre-trained decoder during fine-tuning and randomly initialize a new task-specific decoder for each new downstream task, causing significant pretrain-finetune gap. 

We demonstrate that we can apply the proposed GLID to 6 computer vision tasks without introducing architecture twists. Through extensive experiments, we show that the GLID method, despite its simplicity, can achieve strong performance on downstream vision tasks, outperforming or matching more specialized models such as Mask2Former~\cite{mask2former}, DETR~\cite{detr}, ViTPose~\cite{vitpose}, and BinsFormer~\cite{binsformer}. Additionally, we demonstrate that GLID is more data-efficient on downstream tasks through the generalist encoder-decoder pre-training.

In summary, our contributions are as follows:
\begin{itemize}
    \item We propose GLID, a generalist encoder-decoder pre-training method for various vision tasks. 
    \item In GLID, both the pre-training and fine-tuning tasks are modeled as ``query-to-answer" problems, thus minimizing the pretrain-finetune architectural gap. 
    
    \item Through extensive experiments, GLID is shown to outperform or match specialized models on 6 vision tasks and is more data-efficient through the generalist encoder-decoder pre-training.
\end{itemize}

\section{Related Works}
\noindent \textbf{Mask Image Modeling (MIM)} is a popular pretext task for visual representation learning that aims to reconstruct the masked tokens from a corrupted input. 
The MAE approach~\cite{mae} proposes an asymmetric encoder-decoder architecture and uses the encoder to process partially visible tokens and predicts the corresponding pixel values at the masked positions with a lightweight decoder.
In general, the central objective of previous work on MIM is to devise an improved architectural design~\cite{mae, cae} or reconstruction target~\cite{mae,simmim,beit,peco,dvae}, with the aim of facilitating the acquisition of a superior representation by the model.
However, a common limitation of these works is that only the pre-trained encoder can be transferred to downstream tasks, while the decoder must be replaced with a more complex, task-specific architecture, leading to a significant pretrain-finetune gap, 

\noindent \textbf{Unified architecture} has been gaining popularity in recent years for solving various vision tasks with one framework. Mask2Former~\cite{mask2former} is an example of such an architecture that can handle three segmentation tasks. OneFormer~\cite{jain2022oneformer} builds on Mask2Former and proposes a task-conditioned joint training strategy to unify image segmentation. Beyond image segmentation, other unified models, such as Unified-IO~\cite{lu2022unified}, UniT~\cite{hu2021unit}, Pix2seq v2~\cite{chen2022unified}, X-Decoder~\cite{xdecoder}, and UViM~\cite{kolesnikov2022uvim}, can jointly learn various computer vision tasks over multiple datasets, including pose estimation, object detection, image generation, etc. However, these models typically need more task-specific data to achieve satisfactory performance.

\noindent \textbf{Decoder pre-training} has been proposed to tackle the problem that the task-specific decoder needs to be trained from scratch on downstream tasks. 
DDeP~\cite{DDeP} proposed a decoder pre-training approach based on denoising, which can be combined with supervised pre-training of the encoder.
In UP-DETR~\cite{updetr}, a patch feature reconstruction branch is proposed for the decoder, which is jointly optimized with patch detection. 
Instead of pre-training the encoder and decoder separately with specially designed pre-training tasks, VADeR~\cite{vader-nips20} and DenseCL~\cite{densecl} propose to pre-train the entire architecture through pixel-level contrastive learning and then transfer to dense prediction tasks. In contrast to those works, GLID pre-trains a generalist encoder-decoder model and can better adapt to downstream tasks with minimal architectural gap. 

\begin{figure*}
    \centering
    \includegraphics[width=0.9\linewidth]{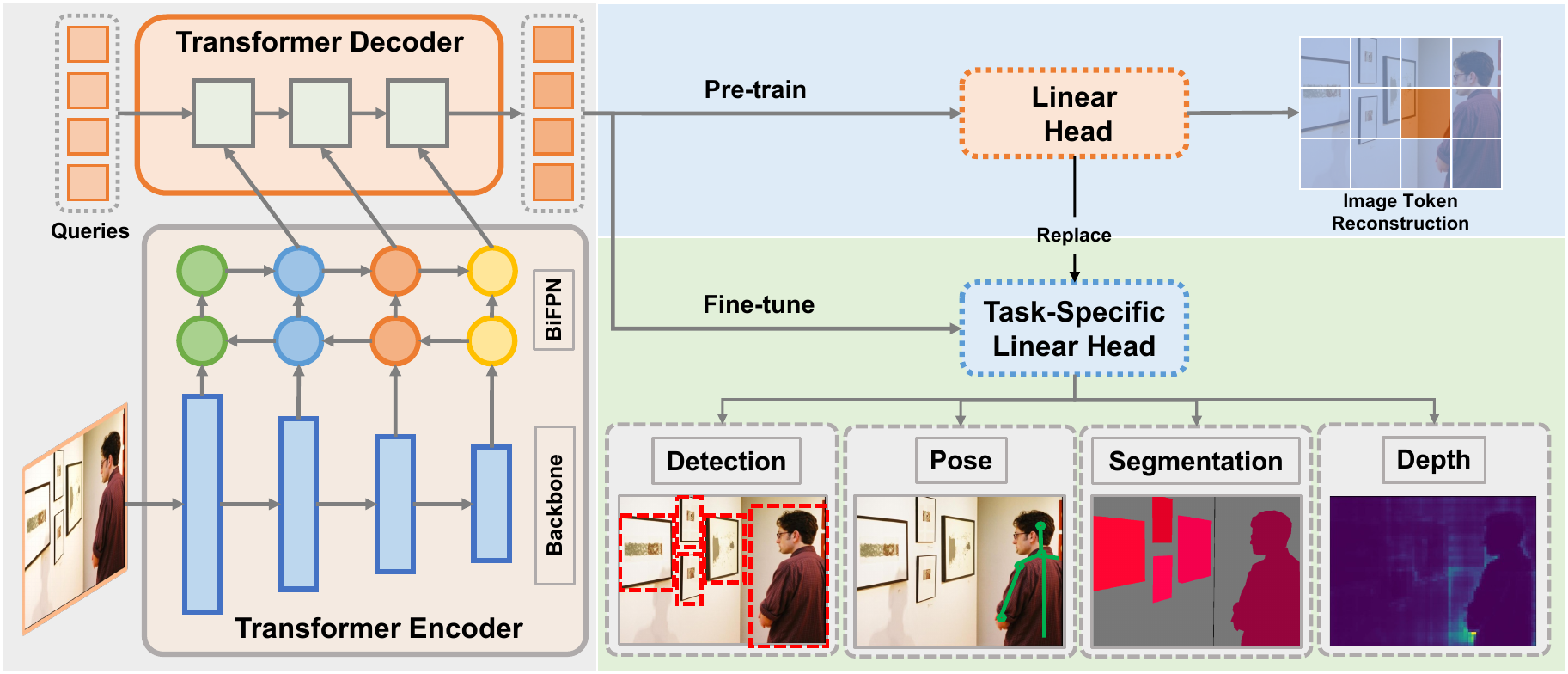}
    \vspace{-1em}
    \caption{Overview of GLID. During pre-training, we pre-train a task-agnostic encoder-decoder transformer architecture through masked image modeling. For fine-tuning on downstream tasks, we replace the pre-training linear head with a task-specific linear head. In this way, the proposed GLID minimizes the pretrain-finetune gap and enables the pre-trained architecture to better adapt to downstream tasks.} 
    \label{fig:GLID}
    \vspace{-1em}
\end{figure*}

\begin{figure}
    \centering
    \includegraphics[width=1.0\linewidth]{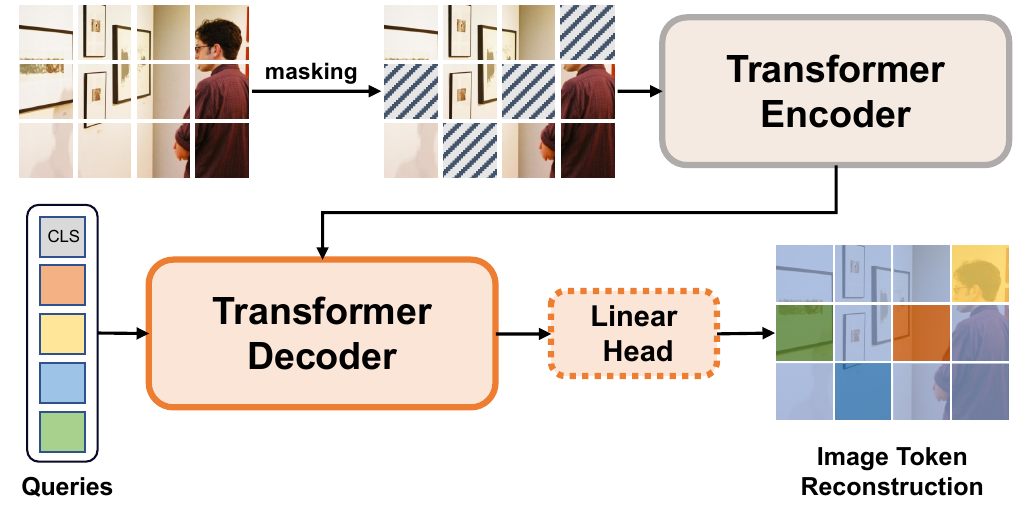}
    \vspace{-2em}
    \caption{GLID pre-training pipeline.} 
    \label{fig:mim}
    \vspace{-1em}
\end{figure}

\section{Methods}

Although pre-training a visual encoder with various strategies has made significant progress, task-specific decoders are still needed for fine-tuning the pre-trained encoder to tackle different downstream tasks~\cite{mask2former, detr, binsformer, vitpose}. 
To tackle this problem, we propose to pre-train a \textbf{G}enera\textbf{LI}st encoder-\textbf{D}ecoder (\textbf{GLID}) vision model for better handling various downstream vision tasks and introduce minimal architecture modifications when shifting from pre-training to fine-tuning. We illustrate the framework of the proposed GLID in \cref{fig:GLID}.

\subsection{Generalist Encoder-Decoder Pre-training}
\label{sec:GLID}
To reduce the pretrain-finetune architectural gap and take full advantage of the pre-training, we pre-train a generalist encoder-decoder architecture that can be directly applied to downstream tasks with a unified formulation for the pre-training and fine-tuning.
Specifically, given an input image $x \in \mathbb{R}^{H \times W \times 3}$ where $H$, $W$ are the image height and width, an image encoder $\mathrm{Enc}$ is used to extract visual features, and a query-based transformer decoder $\mathrm{Dec}$ to decode out the final features, which are converted to the target predictions $P$ by a linear head $H$, 
\begin{equation}
\label{eq:loss}
    P = \mathrm{H}(\mathrm{Dec}(Q, \mathrm{Enc}(x))),
\end{equation}
where $Q$ denotes the queries. All tasks are formulated in this way, including the pre-training and fine-tuning tasks. We use different $Q$-$P$ pairs to switch between different tasks,

We pre-train the encoder-decoder through a specially designed Masked Image Modeling (MIM)~\cite{mae,simmim} pretext task. Specifically, given the input image, we partition it into image patches and convert them into image tokens, and a random mask $M$ is applied to mask out a proportion of input tokens. Then the visual encoder is utilized to only process the visible tokens following MAE~\cite{mae}. 
To align the decoder architecture with the decoders used in downstream tasks, we reformulate the MIM pretext task as a ``query-to-answer"~\cite{detr} problem and use the query-based transformer decoder~\cite{attention} to decode out the masked tokens for pre-training. 
In particular, we introduce a number of mask tokens $\mathrm{[M]}$ as the input queries of the decoder, which are added with different positional embedding to indicate a unique masked position.
Moreover, we append one additional $\mathrm{[CLS]}$ token query~\cite{bert} at the 0th position, which is not associated with a specific masked location but is used to capture global representations. So during pre-training, we can express $Q$ as $Q = [\mathrm{CLS, M_1, M_2,...,M_N}]$ where $N$ denotes the number of the masked tokens. 
After decoding, a linear transformation head is utilized to convert the decoded query representations to target predictions. We use pixel values $P^t$ at the masked locations as the reconstruction targets, so $P$ presents the predicted pixel values in the pre-training stage. We calculate the mean squared error (MSE) loss between $P$ and $P^t$ for pre-training, 
\begin{equation}
\label{eq:loss}
    \mathcal{L}_\mathnormal{rec} = \frac{1}{N} \sum\limits_{i=1}^N {\Vert p_i - p^t_i \Vert_2^2},
\end{equation}
where $p_i$ and $p^t_i$ indicate the reconstructed and original pixel values of $i$th masked tokens. Note that the $\mathrm{[CLS]}$ token is not used to calculate the reconstruction loss, but can be optimized during pre-training because the other queries' predictions are conditioned on the $\mathrm{[CLS]}$ token. We illustrate the pre-training pipeline of GLID in \cref{fig:mim}.

Our formulation enables the pre-training of a generalist encoder-decoder model, which can be directly fine-tuned on downstream tasks without adding complex task-specific decoders, leading to a minimal pretrain-finetune architectural gap. 
In contrast, existing pre-training approaches such as MAE-based~\cite{mae} methods discard their decoders after the pre-training and add a new task-specific decoder for each downstream task's fine-tuning. Consequently, such methods require training extra task-specific decoders on downstream tasks from scratch and cannot take full advantage of the pre-training as the decoders are randomly initialized.

\subsection{Fine-tuning on Downstream Tasks}
\label{sec:fine-tune}
Based on the generalist design, we can apply the pre-trained GLID model to various downstream tasks, including object detection, image segmentation, pose estimation, and depth estimation. Following the formulation in \cref{sec:GLID}, all those tasks are also modeled as new ``query-to-answer"  problems. We can fine-tune GLID to those tasks seamlessly by replacing the topmost linear layer used in pre-training with a task-specific linear layer and assigning different learning targets to the query-answer pairs. We use $M$ to denote the number of queries for downstream tasks, which may be different for different downstream tasks. 
To better utilize the pre-trained weights, we initialize the $M$ queries with the pre-trained $\mathrm{[CLS]}$ token by repeating the $\mathrm{[CLS]}$ token for $M$ times and adding $M$ learnable embeddings to the repeated tokens,
\begin{equation}
\label{eq:loss}
    q_i = [\mathrm{CLS}] + e_i,
\end{equation}
where $q_i$ and $e_i$ denote the $i$th query and $i$th learnable embeddings respectively.  
Note that the learnable embeddings are zero-initialized to be compatible with the pre-trained weights.

\noindent\textbf{Object detection.}\quad
For the object detection task, each query is used to represent an object instance following DETR~\cite{detr}. Two linear layers are utilized to convert the decoded hidden features to a bounding box and class probabilities respectively, $P=\{[bbox_i, cls_i]\}_{i=1}^M$.

\noindent\textbf{Image segmentation.}\quad
Following Mask2Former~\cite{mask2former}, each query is utilized to predict a $C$-dimensional mask embedding and its class,  $P=\{[f^s_i, cls_i]\}_{i=1}^M$, where $f_i^s \in \mathbb{R}^{C}$ and $cls_i$ represent the $i$th mask embedding and its class. 
Thus we can obtain the binary mask predictions via a dot product between the mask embeddings and the backbone's 1/4-scale feature map $f^b \in \mathbb{R}^{\frac{H}{4} \times \frac{W}{4} \times C}$ followed by a sigmoid activation,
\begin{equation}
\label{eq:loss}
    S \in \mathbb{R}^{\frac{H}{4} \times \frac{W}{4} \times M} = \mathrm{sigmoid}(\langle f^b,  f^s \rangle),
\end{equation}
where $S$ denotes the binary mask prediction.

\noindent\textbf{Depth estimation.}\quad
Following AdaBins~\cite{bhat2021adabins} and BinsFormer~\cite{binsformer}, the depth regression task is formulated as a ``classification-regression" problem where the continuous predictions are obtained via a linear combination of bin centers. Specifically, the queries' hidden features are converted to bin lengths and bin embeddings by the depth head, $P=\{[l_i, z_i]\}_{i=1}^M$ where $l_i \in \mathbb{R}$, $z_i \in \mathbb{R}^{C}$ are the length and embedding of the $i$th bin. We follow the pose process in AdaBins to obtain the depth maps. We first obtain the bins' centers as:
\begin{equation}
    c(l_i) = d_{min} + \left(d_{max}-d_{min}\right)\left(\frac{l_i}{2}+\sum\limits_{j=1}^{i-1}l_j\right),
\end{equation}
\noindent where $c(l_i)$, $d_{max}$, and $d_{min}$ are the center depth of the $i$th bins, maximum valid depth, and minimum valid depth, respectively. Then we can obtain a probability distribution map $P^d$ via a dot product between the bins embeddings and the backbone’s 1/4-scale feature map followed by a softmax function,
\begin{equation}
\label{eq:loss}
    P^d \in \mathbb{R}^{\frac{H}{4} \times \frac{W}{4} \times M} = \mathrm{softmax}(\langle f^b, z\rangle).
\end{equation}
where $f^b \in \mathbb{R}^{\frac{H}{4} \times \frac{W}{4} \times C}$ denotes the backbone's feature map. Finally, the depth map can be obtained via a linear combination of bin centers, 
\begin{equation}
    d = \sum\limits_{i=1}^{M}c(l_i)p^d_i,
\end{equation}
where $p^d_i \in \mathbb{R}^{\frac{H}{4} \times \frac{W}{4}}$ denotes the $i$th map of $P^d$.

\noindent\textbf{Pose estimation.}\quad
We utilize a heatmap-based approach for pose estimation following ViTPose~\cite{vitpose}. In particular, each query is used to output a heatmap of a key point. A pose prediction head is employed to convert the query features to $C$-dimensional feature vectors for representing the different key points' heatmaps $P=\{f^k_i\}_{i=1}^M$ with $f_i^k \in \mathbb{R}^{C}$ denoting the $i$th heatmap feature vector. The pose heatmaps can be obtained as
\begin{equation}
\label{eq:loss}
     P^k \in \mathbb{R}^{\frac{H}{4} \times \frac{W}{4} \times M} = \langle f^b, f^k\rangle.
\end{equation}
where $f^b \in \mathbb{R}^{\frac{H}{4} \times \frac{W}{4} \times C}$ denotes the backbone's feature map.

After obtaining the queries' predictions, we follow prior arts~\cite{vitpose,binsformer,detr,mask2former} to use the task loss between the predictions and ground truths for fine-tuning. In particular, we calculate the bipartite matching loss for object detection and image segmentation tasks as for those two tasks, the number of predictions is typically larger than the number of ground truths. For the depth estimation and pose estimation tasks, we use Scale Invariant (SI) regression loss~\cite{eigen2014depth} and smooth $L_1$ loss~\cite{fastrcnn} respectively. 
When fine-tuning GLID, 
we inherit majority of the pre-trained architecture weights to take full advantage of the self-supervised pre-training on large-scale dataset. 

\noindent\textbf{Architecture.}\quad
As we aim at applying GLID to various visual tasks, a hierarchical transformer architecture is utilized for better transferability on downstream tasks. In particular, we use Swin Transformer~\cite{swin} as the visual encoder to produce multi-scale feature maps. A Bi-FPN~\cite{fpn} architecture is utilized for better modeling the interaction across different feature map scales. 
The multi-scale feature maps are then inputted to the decoder of $L$ cross-attention and self-attention blocks to produce final predictions. 
Following Mask2Former~\cite{mask2former}, the decoder transformer processes the multi-scale feature maps with cross-attention layers successively.

\begin{table*}[t]
    \centering
    \resizebox{1.0\linewidth}{!}{
    \begin{tabular}{lc|c|cc|cc|ccc|cc}
    \toprule 
    \multirow{2}{*}{\textbf{Method}} & \multirow{2}{*}{\textbf{Type}}  & \multirow{2}{*}{\textbf{Backbone}} &  \multirow{2}{*}{\textbf{Data}} & \multirow{2}{*}{\textbf{w/ Labels}} &  \multicolumn{2}{c|}{\textbf{COCO}} & \multicolumn{3}{c|}{\textbf{ADE20K}} & \multicolumn{2}{c}{\textbf{NYUv2}}  \\
       & & & & & AP\textsuperscript{kp}$\uparrow$ & AP\textsuperscript{box}$\uparrow$ & mIoU$\uparrow$ & mAP$\uparrow$ & PQ$\uparrow$ & REL$\downarrow$ & RMSE$\downarrow$  \\
    \midrule
    Mask-RCNN~\cite{maskrcnn} & \multirow{8}{*}{Specialist} & ResNeXt-101 & \multirow{7}{*}{ImageNet-1K} & \cmark & 63.1 & 37.1 &  - & - & - & - & -  \\
    Mask2Former~\cite{mask2former} &  & Swin-B &  & \cmark & - & - &  52.4 & 30.7\textsuperscript{$\dagger$} & 44.3\textsuperscript{$\dagger$} & - & -  \\
    ViTPose~\cite{vitpose} &  & ViT-B &  & \xmark & 75.8 & - &  - & - & - & - & - \\
    ViTPose~\cite{vitpose} &  & ViT-L &  & \xmark & 78.3 & - &  - & - & - & - & - \\
    DETR~\cite{detr} &  &  ResNet101-DCN &   & \cmark & - & 44.9  & - & - & $\star$ & - & - \\
    Deformable-DETR~\cite{deformabledetr} &  &  ResNet50-DC5 &   & \cmark & - & 46.9  & - & - & - & - & - \\
    BinsFormer~\cite{binsformer} &  & Swin-B &  & \cmark & - & -  &  - & - & - & 0.104 & 0.362\\
    BinsFormer~\cite{binsformer} &  & Swin-L & ImageNet-21K & \cmark & - & -  &  - & - & - & 0.094 & 0.330\\
    \midrule
    Pix2seq v2~\cite{chen2021pix2seq}  & \multirow{6}{*}{Generalist} & ViT-B & Objects365 & \cmark & 68.0 & 46.5  &  - & $\star$ & $\star$ & - & -\\
    UniT~\cite{hu2021unit} &  & ResNet50 &  COCO, VG, VQAv2 & \cmark & - & 40.8  & - & - & - & - & -\\
    UViM~\cite{kolesnikov2022uvim} &  & ViT-L & ImageNet-21K & \cmark & - & -  &  $\star$ & $\star$ & $\star$ & $\star$ & 0.467\\
    Unified-IO~\cite{lu2022unified} &  & Unified-IO-B & 95 datasets & \cmark& $\star$ & $\star$  &  - & $\star$ & - & $\star$ & 0.469 \\
    Unified-IO~\cite{lu2022unified} &  & Unified-IO-L & 95 datasets & \cmark & $\star$ & $\star$  &  - & $\star$ & - & $\star$ & 0.402\\
    Painter~\cite{painter} &  & ViT-L & ImageNet-1K & \xmark & 72.1 & -  &  49.9 & $\star$ & $\star$ & \textbf{0.080} & 0.288 \\
    \midrule
    \rowcolor[gray]{.9} 
    GLID &  & Swin-B & ImageNet-1K & \xmark & 76.7 & 51.2  &  52.7 & 30.9 & 44.7 & 0.085 & 0.293 \\
    \rowcolor[gray]{.9} 
    GLID &  & Swin-L & ImageNet-1K & \xmark & \textbf{78.5} & \textbf{52.4}  & \textbf{53.9} & \textbf{32.5} & \textbf{45.7} & 0.081 & \textbf{0.287}\\
    \bottomrule
    \end{tabular} 
    }
    \vspace{-1em}
    \caption{Comparison with previous specialized and generalized approaches. The ``Data" denotes pre-training data. The `w/ Labels" denotes the pre-training uses supervised supervision. ``$\star$" denotes the model can perform such task but does not have reported results. ``-" denotes the model cannot perform such task. $\dagger$ denotes our implementation with the official code.}
    \label{tab:sota}
    \vspace{-0.5em}
\end{table*}

\begin{table*}[t]
    \centering
    \resizebox{0.66\linewidth}{!}{
    \begin{tabular}{l|c|cc|ccc|cc}
    \toprule 
    \multirow{2}{*}{\textbf{Pre-train Method}} & \multirow{2}{*}{\textbf{w/ Labels}} & \multicolumn{2}{c|}{\textbf{COCO}} & \multicolumn{3}{c|}{\textbf{ADE20K}} & \multicolumn{2}{c}{\textbf{NYUv2}}  \\
      &  & AP\textsuperscript{kp}$\uparrow$ & AP\textsuperscript{box}$\uparrow$ & mIoU$\uparrow$ & mAP$\uparrow$ & PQ$\uparrow$ & REL$\downarrow$ & RMSE$\downarrow$  \\
    \midrule
    Supervised  & \cmark& 72.1 & 44.8 & 50.0 & 26.2 & 40.7 & 0.182 & 0.510  \\
    EsViT~\cite{esvit} & \xmark& 73.1 & 45.0 & 48.7 & 27.0 & 41.5 & 0.160 & 0.415  \\
    MAE~\cite{esvit} & \xmark& 74.2 & 49.2 & 51.5 & 26.5 & 40.3 & 0.165 & 0.340  \\
    MixMAE~\cite{mixmim} & \xmark& 75.8 & 47.5 & 49.7 & 29.7 & 43.0 & 0.106 & 0.327  \\
    SimMIM~\cite{esvit} & \xmark& 75.6 & 48.9 & 50.6 & 25.9 & 39.2 & 0.098 & 0.343  \\
    \rowcolor[gray]{.9} 
    GLID & \xmark & \textbf{76.7} & \textbf{51.2} & \textbf{52.7} & \textbf{30.9} & \textbf{44.7} & \textbf{0.085} & \textbf{0.293}  \\
    \bottomrule
    \end{tabular}
    }
    \vspace{-1em}
    \caption{Comparison with previous pre-training methods on studied tasks. All entries use Swin-B backbone and are pre-trained on ImageNet-1K. The `w/ Labels" denotes pre-training with classification labels of ImageNet-1K.}
    \label{tab:pre-train}
    \vspace{-1em}
\end{table*}

\section{Experiment Setups}
\label{sec:setup}
We validate the effectiveness of GLID with experiments on various vision tasks.
We first pre-train a generalist encoder-decoder on the ImageNet-1K~\cite{imagenet} dataset in a self-supervised manner. We then fine-tune the pre-trained encoder-decoder to various downstream tasks, including object detection/pose estimation on COCO~\cite{coco}, semantic/instance/panoptic segmentation on ADE20K~\cite{ade20k}, and depth estimation on NYU-Depth-v2~\cite{nyu}. 

\noindent\textbf{Dataset and evaluation metrics.}
We use the large-scale ImageNet-1K~\cite{imagenet} dataset for the self-supervised pre-training, which contains 1.2M images. After pre-training, we transfer the pre-trained models to various downstream tasks to demonstrate the transferability of our pre-training. In particular, we transfer to the ADE20K~\cite{ade20k} dataset to conduct fine-tuning on the semantic, instance, and panoptic segmentation tasks. For those three tasks, we use the mean intersection over union (mIoU), mean average precision (mAP), and panoptic quality (PQ)~\cite{kirillov2019panoptic} metrics for performance comparison. We also fine-tune the pre-trained models to the depth estimation task on the NYU-Depth V2~\cite{nyu} dataset. By default, we report the root mean square error (RMSE) and mean absolute relative error (REL) for performance evaluation. Moreover, we also transfer to the large-scale COCO~\cite{coco} dataset and conduct fine-tuning on the pose estimation task and object detection task, and report the mean average precision of pose estimation AP\textsuperscript{kp} and object detection AP\textsuperscript{box} respectively for performance evaluation.

\noindent\textbf{Pre-training.} 
By default, we pre-train for 600 epochs with the input size of $ 192\times192$ and batch size 2048. The Swin-B and Swin-L~\cite{swin} are utilized as the backbones. We use 6 transformer blocks in the decoder with a hidden embedding dimension of 512. We randomly mask the input images with a mask ratio of 75\%. The AdamW~\cite{adamw} optimizer is utilized with an initial learning rate of $1.2 \times 10^{-3}$ and weight decay of 0.05. We apply random resizing cropping and horizontal flipping for data augmentation following MAE~\cite{mae}.

\noindent\textbf{Fine-tuning.} We conduct fine-tuning experiments on various downstream tasks following previous practice. Note that during fine-tuning, we do not introduce architecture modifications except for task-specific linear head, thus can inherit most of the pre-trained weights. In particular, for the three segmentation tasks, we fine-tune for 160K iterations with batch size of 16 and input size of $640 \times 640$. Following Mask2Former~\cite{mask2former}, we use 100/200/200 queries for semantic/instance/panoptic segmentation. 
For depth estimation, we use official data split and fine-tune for 38.4K iterations with batch size of 16. We use 64 queries following the practice in BinsFormer~\cite{binsformer}.
For object detection, we fine-tune for 50 epochs with batch size of 16. Following H-DETR~\cite{hdetr}, we use 1,800 queries for training and 300 queries for evaluation. 
For pose estimation, we train and evaluate on the cropped regions which contain person instances. We fine-tune for 210 epochs with 17 queries.

\noindent\textbf{Multitask fine-tuning.} We also explore fine-tuning with multiple tasks simultaneously. We use the ADE20K~\cite{ade20k} dataset and conduct fine-tuning with different segmentation tasks jointly. Except for the task-specific linear head, other parameters are shared. During fine-tuning, the batch samples of different tasks are evenly included. The training hyper-parameters are the same as those used for single-task fine-tuning.

\section{Main Results}
In this section, we compare our GLID to prior arts on various benchmarks. We first conduct comparisons between GLID and previous pre-training approaches in \cref{sec:exp:pre-train}. We then compare our results with previous generalized and specialized models in \cref{sec:exp:sota}. Lastly, we demonstrate the multi-task learning results on segmentation tasks in \cref{sec:exp:multitask}.

\subsection{Comparison with pre-training methods}
\label{sec:exp:pre-train}
We compare our pre-training method, GLID, with previous pre-training approaches to demonstrate its effectiveness on various visual tasks. Several pre-training approaches are utilized, including the supervised pre-training on ImageNet-1K, the contrastive approach EsViT~\cite{esvit}, and the masked-image-modeling approaches~\cite{simmim,mae,mixmim}. Using the pre-trained backbone of the same architecture, we fine-tune them on the studied downstream tasks and compare their performances in \cref{tab:pre-train}. Our approach outperforms other compared approaches in terms of all reported metrics, demonstrating the effectiveness of our pre-training method. 

Particularly, on the three tasks of the ADE20K~\cite{ade20k} dataset, we improve the supervised pre-training for an average of 3.8 points. Compared to the recent self-supervised learning methods MixMAE~\cite{mixmim} and EsViT, we achieve an average of 1.9 and 3.7 points improvements on these three segmentation tasks. On the NUYv2~\cite{nyu} depth estimation task, GLID achieves 0.293 RMSE (root mean square error), 0.29 better than the supervised pre-training. In addition, compared to the self-supervised EsViT, our GLID can improve its performance by 0.222 RMSE. On the larger-scale COCO~\cite{coco} dataset, GLID can still outperform other pre-training approaches. In particular, we improve the supervised pre-training by 4.6 and 6.4 points on AP\textsuperscript{kp} and AP\textsuperscript{box} respectively. Compared to the self-supervised pre-training MixMAE, we surpass its performance by 0.9 AP\textsuperscript{kp} and 3.7 AP\textsuperscript{box}. All the results demonstrate the effectiveness of our generalist encoder-decoder pre-training.

\subsection{Comparison with specialists and generalists}
\label{sec:exp:sota}

\Cref{tab:sota} shows comparisons of our approach with other specialized and generalized approaches. Our generalist approach can tackle the studied tasks with minimal architectural modifications and achieves strong performance on those tasks. In particular, we only use the ImageNet-1K for self-supervised pre-training, while other approaches use a comparable or stronger setting than ours, such as more pre-training data or supervision.

Specifically, we achieve better performance than Mask2Former~\cite{mask2former} on the three segmentation tasks with the same backbone. On the depth estimation task, our large model achieves 0.287 RMSE, demonstrating a significant improvement on the previous state-of-the-art BinsFormer~\cite{binsformer}. On the COCO keypoint estimation and object detection task, our approach can surpass the specialized approaches ViTPose~\cite{vitpose} and DETR~\cite{detr}. Additionally, compared to the generalized approaches which typically use a larger-scale dataset for pre-training, our approach can still outperform their results by using less pre-training data. In particular, we improve the recent Pix2seq v2~\cite{chen2021pix2seq} on COCO by 10.5 AP\textsuperscript{kp} and 5.9 AP\textsuperscript{box}. While the generalized approaches typically fail to improve the specialized models, our approach achieves strong performance and can match or outperform the specialized approaches, demonstrating the effectiveness of our method. 

In addition, our pre-training method can be scaled up to large models. In particular, our pre-trained Swin-L~\cite{swin} outperforms Swin-B on all studied tasks. For the three segmentation tasks on ADE20K, the large model has an average of 1.2 mIOU improvements over the base model. On the COCO dataset, the large model improves the base model by 1.8 AP\textsuperscript{kp} and 1.2 AP\textsuperscript{box}.

\subsection{Results of multitask fine-tuning}
\label{sec:exp:multitask}

Apart from separately fine-tuning the pre-trained model on various tasks, we also explore the multitask fine-tuning. Specifically, we employ the ADE20K dataset for fine-tuning as its annotations can be used for three types of segmentation tasks. We fine-tune the pre-trained Swin-B on this dataset with different task combinations. 

As shown in \cref{tab:multitask}, we find that the multitask models underperform the single-task models. In particular, under the multitask setting, the best performances for the three tasks are 52.1 mIoU, 29.1 mAP, and 43.2 PQ, which are 0.6, 1.8, and 1.5 worse than the single-task performances. The results are may be caused by task interference. Specifically, compared to training semantic and panoptic tasks, training all three tasks results in 0.3 mIoU and 1.1 PQ drops. However, the results also demonstrate that adding more tasks can also lead to performance improvements. In particular, compared to training instance and panoptic tasks, adding semantic task improves the mAP and PQ for 0.5 and 0.6 points respectively.

\begin{table}[t]
    \centering
    \resizebox{0.9\linewidth}{!}{
    \begin{tabular}{ccc|ccc}
    \toprule 
    \textbf{Semantic} & \textbf{Instance} & \textbf{Panoptic} & \textbf{mIoU} & \textbf{mAP} & \textbf{PQ} \\
    \midrule
    \multicolumn{3}{c|}{Single task} & \textbf{52.7} & \textbf{30.9} & \textbf{44.7} \\
    \midrule
    \cmark & \cmark & \textcolor{gray}{\xmark} & 51.3 & 29.1 & - \\
    \cmark & \textcolor{gray}{\xmark} & \cmark & 52.1 & - & 43.2 \\
    \textcolor{gray}{\xmark} & \cmark & \cmark & - & 28.0 & 41.5 \\
    \cmark & \cmark & \cmark & 51.8 & 28.5 & 42.1 \\
    \bottomrule
    \end{tabular}
    }
    \vspace{-1em}
    \caption{Multitask \textit{vs.} single task fine-tuning on ADE20K.}
    \label{tab:multitask}
    \vspace{-1em}
\end{table}

\section{Ablation Studies}
In this section, we conduct ablation studies to evaluate the impact of different design choices of our proposed \textbf{G}enera\textbf{LI}st encoder-\textbf{G}ecoder pre-training (\textbf{GLID}) on the final performance. Unless otherwise specified, we use the Swin-B~\cite{swin} as the encoder and 6 transformer blocks in the decoder and pre-train the encoder-decoder architecture for 600 epochs on ImageNet-1K~\cite{imagenet}. We fine-tune the pre-trained models on the ADE20K~\cite{ade20k} semantic segmentation task and NYUv2~\cite{nyu} depth estimation task. For performance comparison, we report mIoU on ADE20K and RMSE on NYUv2.

\begin{table}[t]
    \centering
    \resizebox{0.8\linewidth}{!}{
    \begin{tabular}{ccc|cc}
    \toprule 
    \textbf{Backbone} & \textbf{BiFPN} & \textbf{Decoder} & \textbf{mIoU$\uparrow$} & \textbf{RMSE$\downarrow$}  \\
    \midrule
    \textcolor{gray}{\xmark} & \textcolor{gray}{\xmark} & \textcolor{gray}{\xmark} & 34.9 & 0.753 \\
    \cmark & \textcolor{gray}{\xmark} & \textcolor{gray}{\xmark} & 50.9 &  0.363 \\
    \cmark & \cmark & \textcolor{gray}{\xmark} & 51.1 & 0.341 \\
    \rowcolor[gray]{.9} 
    \cmark & \cmark & \cmark & \textbf{52.7} & \textbf{0.293} \\
    \bottomrule
    \end{tabular}
    }
    \vspace{-1em}
    \caption{Ablation of loading partial pre-trained weights. The results in row 1 are from scratched training.}
    \vspace{-1em}
    \label{tab:abs:load}
\end{table}

\begin{table}[t]
    \centering
    \resizebox{0.62\linewidth}{!}{
    \begin{tabular}{c|cc}
    \toprule 
     \textbf{\# decoder layers} & \textbf{mIoU$\uparrow$} &\textbf{ RMSE$\downarrow$} \\
    \midrule
    3 & 50.7 & 0.302 \\
    \rowcolor[gray]{.9} 
    6 & \textbf{52.7} & 0.293 \\
    9 & \textbf{52.7} & \textbf{0.292} \\
    \bottomrule
    \end{tabular}
    }
    \vspace{-1em}
    \caption{Ablation of the number of decoder layers.}
    \vspace{-1em}
    \label{tab:aba:decoder}
\end{table}

\noindent\textbf{Load partial pre-trained weights.}
To investigate the importance of our generalist encoder-decoder pre-training, we conducted an ablation study by loading only partial pre-trained weights for fine-tuning. For the unloaded architecture, we initialize it with random weights. We then fine-tune the model on two downstream tasks and compared their performance with the full pre-trained model.

As shown in \cref{tab:abs:load}, the performance gradually improves as more pre-trained weights are loaded. In particular, compared to only loading the backbone weights as in previous works, loading all pre-trained weights improves the mIoU by 1.8 points and reduces the RMSE by 0.07. Additionally, loading the BiFPN part can also outperform the choice of only loading the backbone weights.
All the results demonstrate the importance of the generalist encoder-decoder pre-training.

\begin{table}[t]
    \centering
    \resizebox{0.7\linewidth}{!}{
    \begin{tabular}{c|c|c}
    \toprule 
    \textbf{ \# pre-training epochs} &\textbf{ mIoU$\uparrow$} &\textbf{ RMSE$\downarrow$} \\
    \midrule
    0 &   34.9  &  0.753  \\
    300 & 51.5  & 0.314 \\
    \rowcolor[gray]{.9} 
    600 & 52.7 & 0.293 \\
    900 & \textbf{52.8} & \textbf{0.290} \\
    \bottomrule
    \end{tabular}
    }
    \vspace{-1em}
    \caption{Ablation of the number of pre-training epochs.}
    \label{tab:aba:epochs}
    \vspace{-1em}
\end{table}

\begin{table}[t]
    \centering
    \resizebox{0.8\linewidth}{!}{
    \begin{tabular}{cc||cc}
    \toprule
    \textbf{Masking strategy} & \textbf{mIoU} & \textbf{Masking ratio} & \textbf{mIoU} \\
    \midrule
    grid & 50.8 & 0.5 & 49.8 \\
    block & 49.3 & 0.6 & 51.0 \\
    \rowcolor[gray]{.9} 
    random & \textbf{51.5} & 0.75 & \textbf{51.5} \\
    \bottomrule
    \end{tabular}
    }
    \vspace{-1em}
    \caption{Ablation of masking strategies and ratios.}
    \vspace{-1em}
    \label{tab:aba:mask}
\end{table}

\begin{table}[t]
    \centering
    \resizebox{0.85\linewidth}{!}{
    \begin{tabular}{cc|c|c}
    \toprule 
     \textbf{\% fine-tuning data} & \textbf{Pre-training} &\textbf{ mIoU$\uparrow$} & \textbf{RMSE$\downarrow$} \\
    \midrule
    \multirow{2}{*}{10} & SL & 27.0 & 0.779 \\
     & GLID & \textbf{31.2} & \textbf{0.317} \\
    \midrule
    \multirow{2}{*}{20} & SL & 31.3 & 0.681 \\
     & GLID & \textbf{35.0} & \textbf{0.303} \\
    \midrule
    \multirow{2}{*}{50} & SL & 43.2 & 0.620  \\
     & GLID & \textbf{46.3} & \textbf{0.295} \\
    \midrule
    \multirow{2}{*}{100} & SL & 50.0 & 0.510 \\
     & GLID & \textbf{52.7} & \textbf{0.293} \\
    \bottomrule
    \end{tabular}
    }
    \vspace{-1em}
    \caption{Fine-tuning with limited data. ``SL" denotes using the supervised pre-training on ImageNet-1K. }
    \label{tab:aba:data}
    \vspace{-1em}
\end{table}

\begin{figure*}
    \centering
    \includegraphics[width=0.75\linewidth]{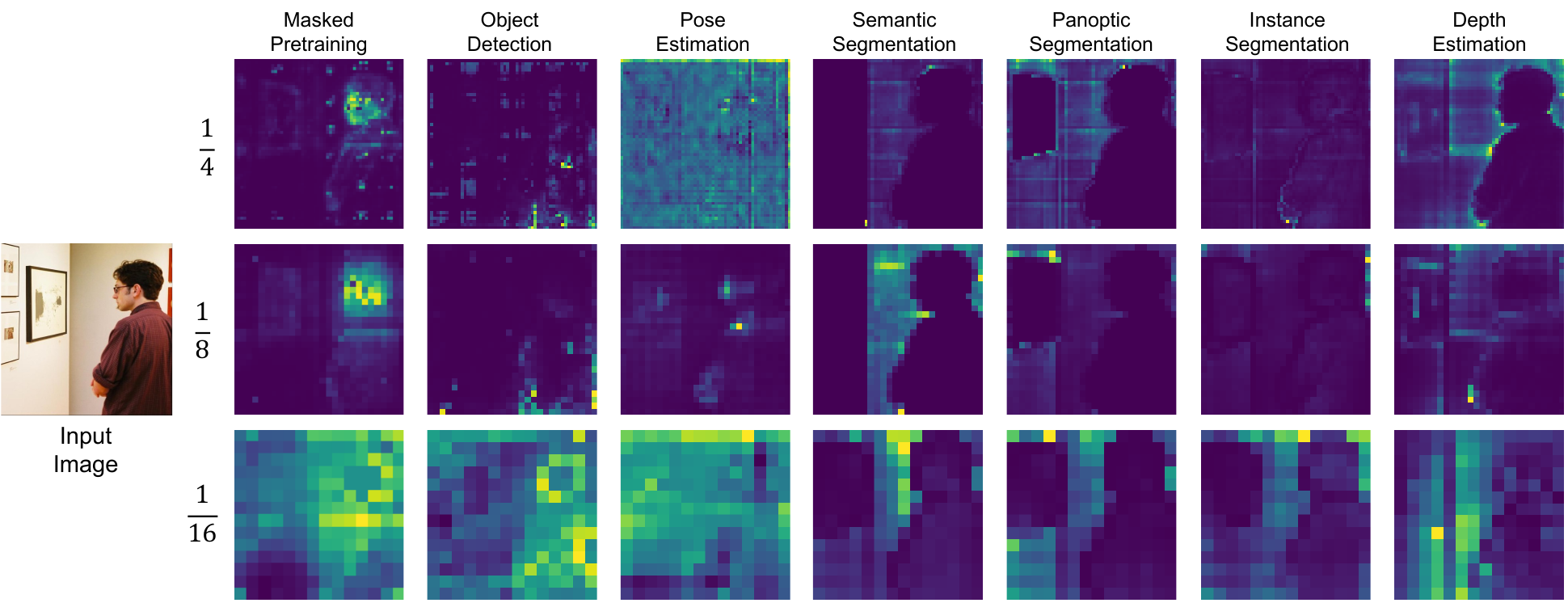}
    \vspace{-1em}
    \caption{Query's cross-attention maps on different tasks. Each column shows three-scale attention maps on a specific task. Task-specific queries are utilized for visualization.}
    \vspace{-1em}
    \label{fig:attn}
\end{figure*}

\begin{figure*}
    \centering
    \includegraphics[width=0.6\linewidth]{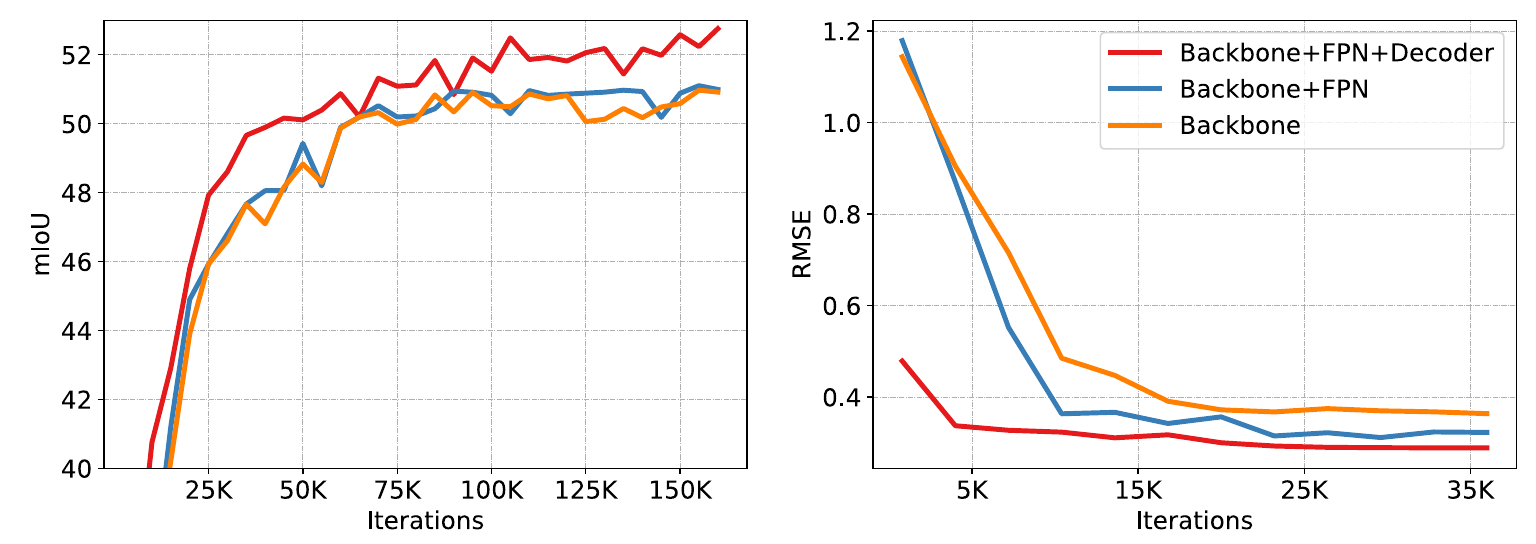}
    \vspace{-1em}
    \caption{Performance curves of loading different parts of pre-trained weights. }
    \vspace{-1em}
    \label{fig:acc_curve}
\end{figure*}
\noindent\textbf{Decoder depth.}
To investigate the impact of decoder depth on our model, we test varying the number of decoder layers in the model. Specifically, we pre-train models with 3, 6, and 9 decoder layers, and fine-tune them on downstream tasks. Note that we use the same number of decoder layers for both pre-training and fine-tuning. As shown in \cref{tab:aba:decoder}, we find that using 3 layers results in 2.0 mIoU and 0.009 RMSE performance drops compared to those of using 6 layers. Additionally, using more decoder layers does not bring significant performance gains but introduces extra computation costs.

\noindent\textbf{Pre-training epochs.} 
To investigate the effect of the number of pre-training epochs on the performance of the model, we vary the number of pre-training epochs. Specifically, we pre-train with different numbers of epochs and fine-tune the pre-trained models on downstream tasks. As shown in \cref{tab:aba:epochs}, we find that we can improve the performance by pre-training more epochs. Specifically, compared to pre-training with 300 epochs, we can improve the mIoU by 1.2 and RMSE by 0.021 with extra 300 epochs of pre-training. However, we find that the performance gradually saturates and does not improve much if we pre-train for 900 epochs.

\noindent\textbf{Mask strategy and mask ratio.} 
We investigate the impact of masking strategy and masking ratio and show the results in \cref{tab:aba:mask}. We pre-train for 300 epochs and report the mIoU on ADE20K for comparison. In particular, we find that using random masking performs best among other masking strategies. Additionally, using a large mask ratio, \ie, 0.75, leads to the best result.

\noindent\textbf{Fine-tuning with limited data.} 
We investigate the impact of limited data during fine-tuning. Specifically, given the GLID and ImageNet-1K-supervised pre-training models, we fine-tune them on downstream tasks with different percentages of data. 
As shown in \cref{tab:aba:data}, we find that our approach has larger performance advantage over the supervised pre-training models when using fewer data for fine-tuning. In particular, when using 10\% fine-tuning data, our approach surpasses the supervised pre-training by 4.2 mIoU and 0.09 RMSE respectively. When we use 100\% data for fine-tuning, the performance gaps are 2.6 mIoU and 0.047 RMSE respectively. The main reason is that for the supervised pre-training approach, we need to train the task-specific decoder from scratch on downstream tasks, 
In comparison, our unified pre-training approach can pre-train and keep pre-trained encoder-decoder simultaneously, and achieve strong performance even without much task-specific data.

\section{Qualitative Evaluation}
\paragraph{Query's cross-attention maps on different tasks.}
We show the query's cross-attention maps on various tasks in \cref{fig:attn}. We observe that the query can focus on different semantic parts for different tasks. In particular, for the pre-training pretext task, the query learns to attend to the local regions for reconstruction. For the object detection and pose estimation tasks, the query attends to some extreme points, such as the boundary of the object or a specific key point location. Moreover, for the segmentation or depth estimation tasks, the attention maps have a clear separation for different semantic regions.

\noindent\textbf{Convergence curves.}\quad
We compare the convergence curves of partially loading the pre-trained weights on the semantic segmentation and depth estimation tasks, with results shown in \cref{fig:acc_curve}. We observe that loading all the pre-trained weights leads to a much faster convergence speed and can improve the final performance. In particular, compared to only loading the backbone's weights, loading all the pre-trained weights can achieve the same performance by using half of the iterations. 
Unlike the traditional approaches~\cite{mixmim,mae} that only load the pre-trained backbone, our generalist pre-training approach allows the pre-trained encoder-decoder to be applied on downstream tasks to obtain better performance.

\section{Conclusion}

In this paper, we propose GLID, a generalist encoder-decoder pre-training method for better handling various downstream computer vision tasks. In the GLID pre-training scheme, both pre-training and fine-tuning tasks are modeled as ``query-to-answer" problems. 
GLID introduces minimal task-specific architecture modifications and can inherit most pre-trained weights when fine-tuning on downstream tasks, which takes full advantage of the pre-training.
We demonstrate that GLID can outperform or achieve comparable performance to specialized models on different vision tasks through extensive experimentation. In addition to better performance, the experimental results indicate that GLID leads to faster convergence speed and data efficiency.

\paragraph{Acknowledgement} This project is funded in part by National Key R\&D Program of China Project 2022ZD0161100, by the Centre for Perceptual and Interactive Intelligence (CPII) Ltd under the Innovation and Technology Commission (ITC)’s InnoHK, by General Research Fund of Hong Kong RGC Project 14204021. Hongsheng Li is a PI of CPII under the InnoHK.

{
    \small
    \bibliographystyle{ieeenat_fullname}
    \bibliography{main}
}

\end{document}